\newif\iffinal
\finaltrue
\documentclass{article}
\usepackage{spconf,amsmath,epsfig}

\usepackage{hyperref}
\usepackage{url}
\usepackage[capitalise]{cleveref}
\usepackage{subcaption}
\usepackage{booktabs}
\usepackage[inline]{enumitem}
\usepackage{multirow}
\usepackage{paralist}
\usepackage{balance}

\title{3D Fully Convolutional Neural Networks\\ With Intersection Over Union Loss\\ for crop mapping from multi-temporal satellite images}
\iffinal
\name{Sina Mohammadi, Mariana Belgiu, Alfred Stein}
\address{Dept. of Earth Observation Science, ITC Faculty, University of Twente,\\ Enschede, The Netherlands\\\{s.mohammadi, m.belgiu, a.stein\}@utwente.nl}
\else
\name{Names}
\address{Draft}
\fi

\begin{document}
%
\maketitle

\begin{abstract}
Information on cultivated crops is relevant for a large number of food security studies. Different scientific efforts are dedicated to generating this information from remote sensing images by means of machine learning methods. Unfortunately, these methods do not take account of the spatial-temporal relationships inherent in remote sensing images. In our paper, we explore the capability of a 3D Fully Convolutional Neural Network (FCN) to map crop types from multi-temporal images. In addition, we propose the Intersection Over Union (IOU) loss function for increasing the overlap between the predicted classes and ground reference data. The proposed method was applied to identify soybean and corn from a study area situated in the US corn belt using multi-temporal Landsat images. The study shows that our method outperforms related methods, obtaining a Kappa coefficient of 91.8\%. We conclude that using the IOU loss function provides a superior choice to learn individual crop types.
\end{abstract}
\begin{keywords}
Crop mapping , deep learning, fully convolutional neural networks , time series.
\end{keywords}
%
\vspace*{-0.7em}
\section{Introduction}
\vspace*{-0.7em}


Multi-temporal remote sensing images are being generated at an unprecedented scale and rate from resources such as Sentinel-2 (5 days frequency), Landsat (16 days frequency), and PlanetScope (daily). In the light of this, there have been many scientific efforts towards converting huge quantities of multi-temporal remote sensing images into useful information. One of these scientific efforts is automatic crop mapping \cite{ji20183d, zhong2019deep, xu2020deepcropmapping, belgiu2021phenology}, being an active research area in remote sensing. 

A decisive factor towards the goal of crop classification from multi-temporal images is developing methods that can learn temporal relationships in image time series. Traditional approaches for temporal feature representation such as  Multi layer Perceptron, Random Forest, Support Vector Machine, and Decision Tree \cite{khatami2016meta, king2017multi,low2013impact,massey2017modis,shi2016assessment} have been developed for single-date images and are not able to explicitly consider the sequential relationship of multi-temporal data.

Recently with the success of deep neural networks in learning high-level task-specific features, CNN and LSTM-based methods have drawn increasing attention and achieved promising results in the field of crop classification from multi-temporal images \cite{xu2020deepcropmapping,zhong2019deep,pelletier2019temporal,danilla2017classification,ji20183d,kussul2017deep}. While most deep learning-based methods for crop mapping use pixel-by-pixel approach, in this paper, we will design a Fully Convolutional Neural Network (FCN) and use it for crop mapping. FCNs have been widely used in semantic segmentation, salient object detection, as well as brain tumor segmentation \cite{yu2018learning,chen2017deeplab,mohammadi2020cagnet, noori2019attention}. They are capable of generating the segmentation map of the whole input image at once and thus are more efficient computationally. In addition, the spatial relationship between adjacent pixels is taken into account by using FCNs in contrast to pixel-by-pixel approaches, which take individual pixels as input.  To fit the need of crop mapping, i.e. learning the sequential relationship of multi-temporal remote sensing data, we use 3D convolution operators as the building blocks of this FCN. They allow both the spatial and the temporal features to be extracted simultaneously. This would be beneficial to crop mapping since both the spatial and temporal relationships in multi-temporal remote sensing data are important for accurate crop mapping.

To learn different crop types, most deep learning-based crop mapping methods use the cross-entropy loss  \cite{xu2020deepcropmapping,danilla2017classification,kussul2017deep,ji20183d,pelletier2019temporal,russwurm2018multi}. They achieved promising results, but we hypothesize that there is still room for improvement by using a loss function better suited for cop mapping than the cross-entropy loss. To guide the network to generate more accurate prediction for crop types, we propose to learn the crop types by increasing the overlap between the prediction map and ground reference mask directly rather than using the cross-entropy loss that only focuses on per-pixel prediction. To the best of our knowledge, this is the first attempt to use this loss function in crop mapping.

In summary, the main contribution of this paper is to learn to identify different crop types by increasing the overlap between the prediction map and ground reference mask for each crop type, which would result in a rethink of the loss functions used to train deep neural networks for crop mapping. In conjunction with this loss function, we design a 3D FCN to simultaneously take into account the spatial and the temporal relationships in multi-temporal remote sensing data.

\begin{figure*}[!t]
    \centering
    \includegraphics[width=14.4cm,height=8.9cm]{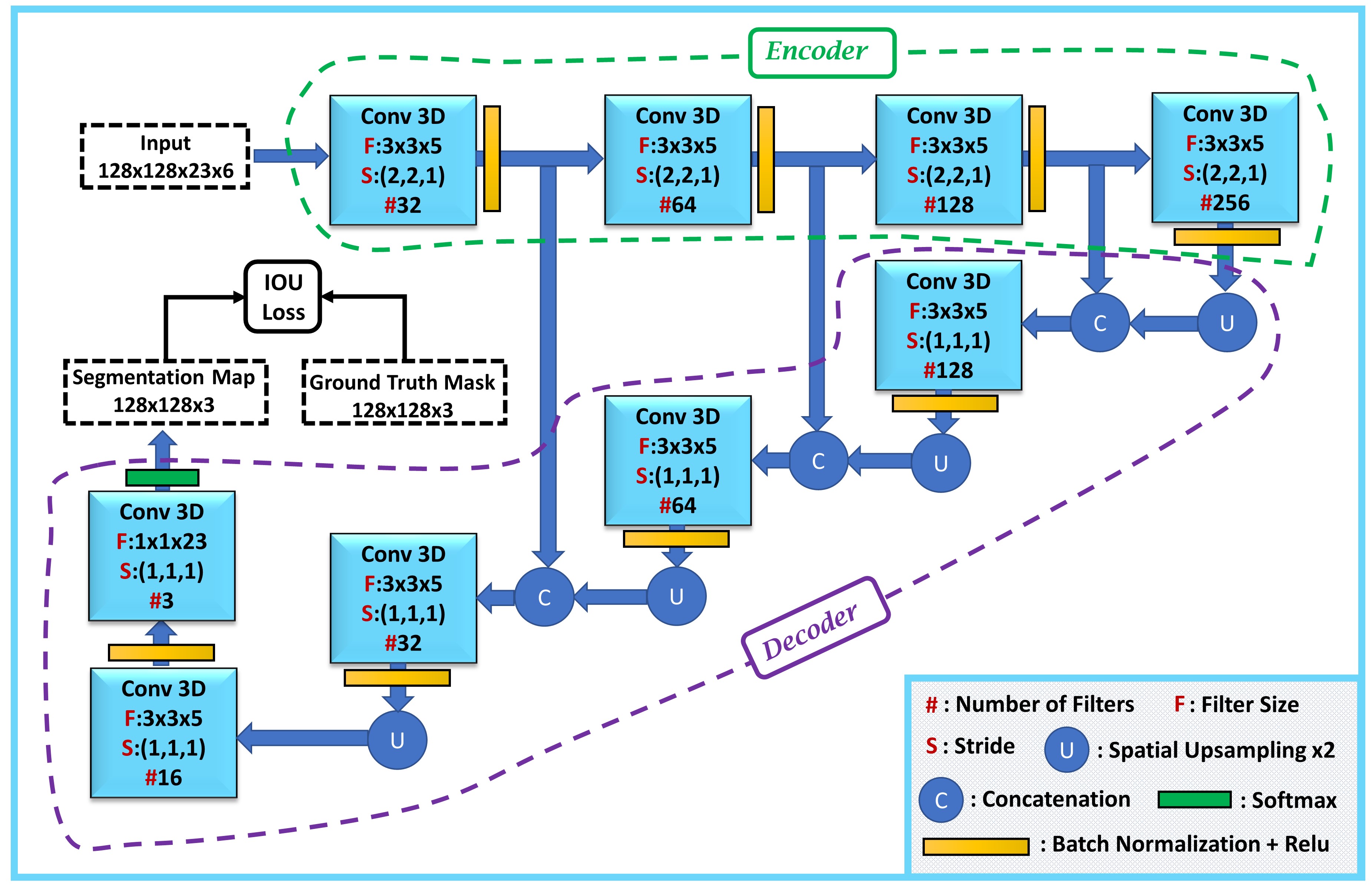}
    \caption{The architecture of the designed 3D FCN, composed of an Encoder and a Decoder, and trained using the IOU loss function.}
    \label{fig:3DFCN}
\end{figure*}

\vspace*{-0.7em}
\section{The proposed method}
\vspace*{-0.7em}

In this section, we explain our designed 3D FCN and Intersection Over Union (IOU) loss function, which is used to train the network. This network (\cref{fig:3DFCN}) is composed of an encoder-decoder network. It learns to generate the segmentation map of crop lands from the input images. 
One important component of our proposed FCN is the 3D convolutional operator that is more beneficial than 2D convolutional operator for multi-temporal crop mapping since it also extracts the temporal features in addition to the spatial features. In the 3D FCN architecture, the Encoder extracts features at four different levels, each of which has different recognition information from each other. At lower levels, the Encoder captures spatial and local information due to the small receptive field, whereas it captures semantic and global information at higher levels because of the large receptive field. To take advantage of the both high level global contexts and low level details, features of different levels are merged in the Decoder through concatenation as shown in \cref{fig:3DFCN}. In conjunction with the 3D FCN, we propose to use Intersection Over Union (IOU) loss to guide the FCN to output accurate segmentation maps.

In contrast to most deep learning-based crop mapping methods that use cross-entropy loss to learn the crop types, we propose a better loss function to guide the network to learn each crop type more accurately. We propose to use a loss function that tries to increase the overlap between the prediction map and ground reference mask directly. This loss function is more suited to crop mapping than the cross-entropy loss that only focuses on per-pixel prediction. Therefore, to increase the overlap between the prediction map and ground reference, we maximize the Intersection Over Union (IOU) for each crop type by adopting the following loss function:

\begin{equation}
L_{IOU}=\sum_{k=1}^{C} (1-IOU_{k})
\end{equation}
where \textit{C} denotes the number of classes, i.e. number of crop types and \textit{$IOU_{k}$} is defined as:
\begin{equation}
IOU_{k}=\frac{1}{M}{\sum_{m=1}^{M}{\frac{\sum_{i=1}^N p^{k}_{i,m} \cdot g^{k}_{i,m}}{\sum_{i=1}^N [p^{k}_{i,m}+ g^{k}_{i,m}- p^{k}_{i,m} \cdot g^{k}_{i,m}]}}}
\end{equation}
where \textit{M}, \textit{N}, \textit{p}, and \textit{g} denote total number of examples, total number of pixels in each example, prediction map, and ground reference mask respectively.

In the Experimental Results section we will show that using this loss function for learning the crop types results in a boost in the performance compared to using the cross-entropy loss. 
\vspace*{-0.7em}
\section{Experiments}
\vspace*{-0.7em}

\begin{figure*}[!t]
    \centering
    \includegraphics[scale=0.57]{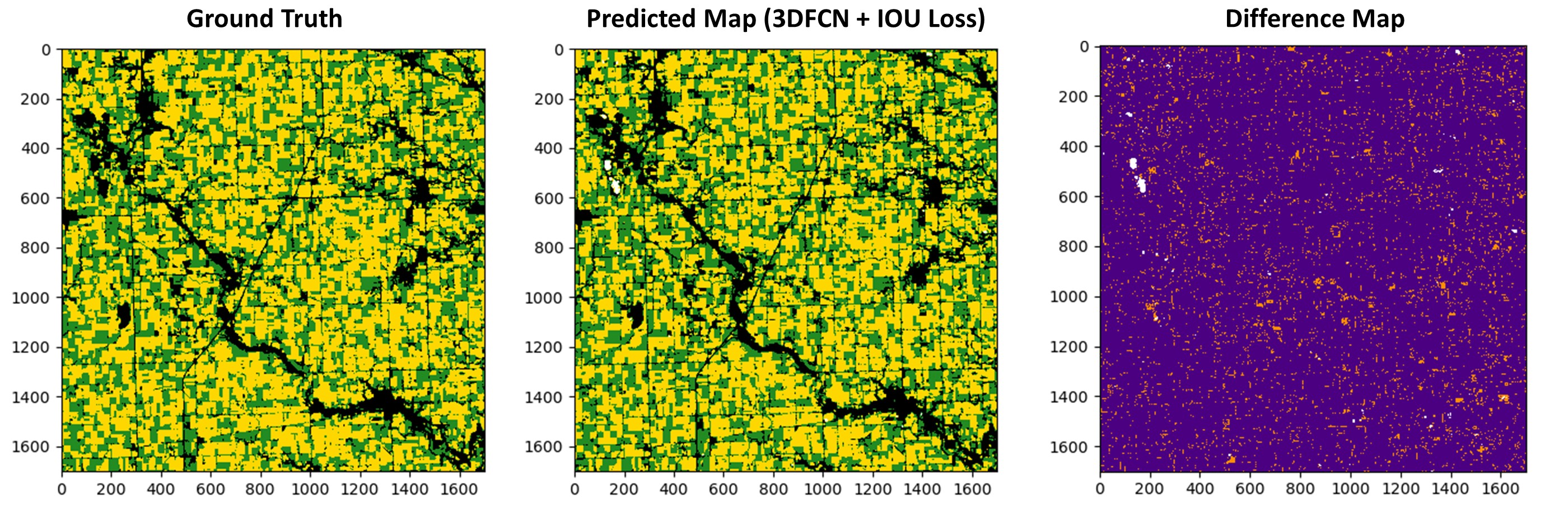}
    \caption{The predicted map of the 3D FCN trained with the IOU loss, ground reference, and difference map. In the figure, green, yellow, and black represent soybean, corn, and other classes respectively. }
    \label{fig:map}
\end{figure*}

\subsection{Study Area, Preprocessing, and Evaluation Metrics}

Our experiments were conducted in the U.S corn belt. We selected a 1700x1700 pixel area located in the center of the footprint of h18v07 in the Analysis Ready Data (ARD) grid system. We used Landsat ARD as the input to our method, which are publicly available on \href{https://earthexplorer.usgs.gov}{USGS's EarthExplorer web portal}. At each observation date, this dataset contains six optical bands, namely red, green, blue, shortwave infrared 1, shortwave infrared 2, and near-infrared. We used \href{https://earthexplorer.usgs.gov}{CropScape website portal} to download the Cropland Data Layer (CDL) as the labels for training, validation, and testing the network. The selected region is mostly composed of corn and soybean. In this project, corn, soybean, and “other” (i.e., merged class of other land cover/land use types) are taken as three classes of interest. Therefore, these three categories are assigned to the corresponding pixels of the input image. We used the Landsat multi-temporal data from April 22 to September 23 that covers growing season of corn and soybean \cite{xu2020deepcropmapping}.

To preprocess the Landsat multi-temporal data and prepare them for training and testing the model, we followed the same procedure as \cite{xu2020deepcropmapping}. We removed the invalid pixels from the dataset. An invalid pixel is a pixel with less than seven valid observations after May 15 \cite{xu2020deepcropmapping}, and a valid observation is the pixel that is not filled, shadowed, cloudy, or unclear. The invalid pixels were excluded from the dataset and were not used in the training process because they do not contain enough multi-temporal information. To fill in the resulted gaps in the valid pixels, linear interpolation was used that resulted in 23 time steps with seven days interval from 22 April to 23 September. Furthermore, we normalized the data using the mean and standard deviation values.

As for performance evaluation of the proposed methods, we employed Cohen's kappa coefficient, macro-averaged producer's accuracy, and macro-averaged user's accuracy \cite{xu2020deepcropmapping}.

\subsection{Implementation details}
We implemented our method in Keras \cite{chollet2015keras} using the Microsoft Azure Machine Learning environment. The designed 3D FCN takes as input a 128x128x23x6 image and outputs a 128x128x3 segmentation map. In the input image size, 128x128, 23, and 6 correspond to the spatial size, number of time steps in time series, and number of optical bands respectively. In the output map size, 128x128 and 3 correspond to spatial size of the segmentation map and number of classes respectively. We used the stochastic gradient descent with a momentum coefficient 0.9 and a learning rate of 0.005. We split the training data into five sections, and used each of them as validation and the rest for training with batch size 2, which resulted in 5 models whose softmax outputs are averaged during testing. The code is publicly available at: \href{https://github.com/Sina-Mohammadi/3DFCNwithIOUlossforCropMapping}{https://github.com/Sina-Mohammadi/3DFCNwithIOUlossforCropMapping}

\begin{table}

    \caption{The experimental results. Kappa, MA-PA, MA-UA, and CE loss stand for Cohen's kappa coefficient, macro-averaged producer's accuracy, macro-averaged user's accuracy, and cross-entropy loss.}
    \centering
    \begin{center}
    \scalebox{0.902}{
    \begin{tabular}{|c|c|c|c|}
    
    \hline
     Method & Kappa& MA-PA & MA-UA\\
    \hline
    Transformer&88.6&90.4&92.1\\
    \hline
    Random Forest&88.7&91.4&91.4\\
    \hline
    Multilayer Perceptron&88.8&91.4&91.5\\
    \hline 
    DeepCropMapping (DCM) \cite{xu2020deepcropmapping} &89.3&91.7&91.9\\
    \hline      
    \textbf{Ours(3DFCN+CE loss)}&\textbf{91.3}&\textbf{93.7}&\textbf{93.6}\\
    \hline
    \textbf{Ours(3DFCN+IOU loss)}&\textbf{91.8}&\textbf{94.1}&\textbf{94.2}\\
    \hline
    
    \end{tabular}}
    \end{center}
     
    \label{tab:results}
\end{table}

\subsection{Experimental Results}
We used the data from the selected study area collected in 2015,2016,2017 as our training set, and we tested the trained 3D FCNs on the data collected in 2018. Then, we compared our method with the baseline classification models, namely Random Forest (RF), Multilayer Perceptron (MLP), and Transformer \cite{vaswani2017attention} with the exact same settings introduced in \cite{xu2020deepcropmapping}. Moreover, we compared our method with the deep learning-based method introduced in \cite{xu2020deepcropmapping}. The results are shown in \cref{tab:results}. As seen from the table, our method outperforms other methods in terms of different evaluation metrics. Moreover, it can be seen that the adopted IOU loss function performs better than the cross-entropy loss. In addition, to visually investigate the performance of the method, we showed the predicted map of the 3D FCN trained with the IOU loss, ground reference, and difference map in \cref{fig:map}.

\vspace*{-0.7em}
\section{Conclusion}
\vspace*{-0.7em}

In this study, a 3D FCN with an IOU loss function has been successfully applied to map soybean and corn crops in the US corn belt. The experimental results show that the adopted IOU loss function, which maximizes the overlap between the prediction map and ground reference mask for each crop type, is able to increase the performance compared to using the widely used cross-entropy loss. Therefore, using the IOU loss function is a better choice to learn individual crop type. For future work, we plan to apply the method to the regions that include more crop types to see to what extent our method can improve the performance.

\vspace*{-0.7em}
\section{Acknowledgement}
\vspace*{-0.7em}

This work was supported by Microsoft under the AI for Earth Microsoft Azure Compute Grant.
\balance
\small{
\bibliography{main}
\bibliographystyle{IEEEbib}
}
\end{document}